\pdfoutput=1
\documentclass[11pt]{article}

\usepackage[]{naacl2021}

\usepackage{times}
\usepackage{latexsym}
\usepackage[utf8]{inputenc} % allow utf-8 input
\usepackage[T1]{fontenc}    % use 8-bit T1 fonts
\usepackage{hyperref}       % hyperlinks
\usepackage{url}            % simple URL typesetting
\usepackage{booktabs}       % professional-quality tables
\usepackage{amsfonts}       % blackboard math symbols
\usepackage{nicefrac}       % compact symbols for 1/2, etc.
\usepackage{enumitem}
\usepackage{graphicx}
\usepackage{color}
\usepackage[protrusion=true,expansion=true]{microtype}

\usepackage[T1]{fontenc}
\usepackage[utf8]{inputenc}
\usepackage{booktabs}
\usepackage{tabularx}

\usepackage{microtype}
\usepackage{graphicx}

\title{Alexa Conversations: An Extensible Data-driven Approach for Building Task-oriented Dialogue Systems}

\author{Anish Acharya\thanks{\enspace Work done while at Amazon}, Suranjit Adhikari,  Sanchit Agarwal,  Vincent Auvray,  Nehal Belgamwar, \\
        {\bf Arijit Biswas, Shubhra Chandra, Tagyoung Chung, Maryam Fazel-Zarandi, Raefer Gabriel,} \\
        {\bf Shuyang Gao,  Rahul Goel\footnotemark[1],  Dilek Hakkani-Tur,  Jan Jezabek,  Abhay Jha,  Jiun-Yu Kao,} \\
        {\bf Prakash Krishnan,  Peter Ku,  Anuj Goyal,  Chien-Wei Lin,  Qing Liu,  Arindam Mandal,} \\
        {\bf Angeliki Metallinou,  Vishal Naik,  Yi Pan,  Shachi Paul\footnotemark[1],  Vittorio Perera, Abhishek Sethi\footnotemark[1],} \\
        {\bf Minmin Shen, Nikko Strom and Eddie Wang}\\
        Amazon Alexa AI, Sunnyvale, California, USA\thanks{\enspace Authors are ordered alphabetically}}

\begin{document}
\maketitle
\begin{abstract}
Traditional goal-oriented dialogue systems rely on various
components such as natural language understanding, dialogue state tracking, policy learning and
response generation. Training each component requires annotations which are
hard to obtain for every new domain, limiting scalability of such systems.
Similarly, rule-based dialogue systems require extensive writing and maintenance of rules
and do not scale either. End-to-End dialogue systems, on the other hand, do not require
module-specific annotations but need a large amount of data for training. To overcome these problems,
in this demo, we present
Alexa Conversations\footnote[1]{\url{https://tinyurl.com/y3lowd34}},
a new approach for building goal-oriented dialogue systems that is scalable, extensible as well as data efficient.
The components of this system are trained in a data-driven manner, but instead of collecting annotated conversations
for training, we generate them using a novel dialogue simulator based on a few seed dialogues and
specifications of APIs and entities provided by the developer. Our approach provides
out-of-the-box support for natural conversational phenomena like entity sharing across turns
or users changing their mind during conversation without requiring developers to provide any such
dialogue flows. We exemplify our approach using a simple pizza ordering task and showcase its
value in reducing the developer burden for creating a robust experience. Finally, we evaluate our
system using a typical movie ticket booking task and show that the
dialogue simulator is an essential component of the system that leads to over $50\%$ improvement
in turn-level action signature prediction accuracy.
\end{abstract}

\section{Introduction}

Goal-oriented dialogue systems enable users to complete specific goals such as
making restaurant reservations and buying train tickets. User goals may be
complex and may require multiple turns to achieve. Moreover, users can refer to contextual values
anaphorically, can correct previously informed preferences and provide additional or fewer
entities (over-cooperative or under-cooperative user) than requested by the agent.
This presents challenges for building robust dialogue agents that need to understand different
kinds of user behavior, gather user requirements split over multiple turns and complete user goals
with minimal friction.
There is also limited availability of dialogue datasets and they span only a
handful of application domains. Designing suitable data collection for dialogue
systems is itself a research area.

Traditional dialogue systems follow a pipelined approach that ties together machine
learning components for natural language understanding (NLU), dialogue state
(belief) tracking, optimal action prediction (policy learning), and natural
language generation~\cite{young2000probabilistic}.
Advances in deep learning techniques have led to the development of more
end-to-end neural dialogue systems that combine some or all of the components of
the traditional pipeline reducing the need for component-wise annotations and
allowing for intermediate representations to be learned and optimized end-to-end
\cite{Wen2017,Liu2017end}. On the data side, notable data collection approaches
for dialogue systems include the Wizard-of-Oz~(WOZ) framework \cite{asri2017frames},
rule-based or data-driven user simulators
\cite{pietquin2005framework,cuayahuitl2005human,pietquin2006probabilistic,schatzmann2007agenda,Fazelzar2017,Gur2018UserMF},
and the recently-proposed Machines-Talking-To-Machines (M2M) framework \cite{Shah2018} where user and
system simulators interact with each other to generate dialogue outlines.

In this demo, we present Alexa Conversations, a novel system that enables developers to build
robust goal-oriented dialogue experiences with minimal effort.
Our approach is example-driven as it learns from a small number of developer-provided seed dialogues
and does not require encoding dialogue flows as rigid rules. Our system contains two core
components: a dialogue simulator that generalizes input examples provided by the developer and a
neural dialogue system that directly predicts the next optimal action given the conversation
history. The dialogue simulator component extends the M2M framework~\cite{Shah2018} in two main
directions. First, instead of generating user goals
randomly, we use various goal sampling techniques biased towards the goals observed in
the seed dialogues in order to support variations of those dialogues robustly. Second, in M2M, the system agent is
geared towards database querying applications where the user browses a catalogue, selects an item and
completes a transaction. In contrast, our formulation does not require any knowledge of the purpose
of the APIs provided by the developer. Moreover, our system can generate a richer set of dialogue patterns including complex goals,
proactive recommendations and users correcting earlier provided entities.
The proposed neural dialogue model component follows an end-to-end systems approach and bears
some similarities with Hybrid Code Networks (HCN) \cite{williams2017hybrid}. However, compared to HCN, our
system is more generic in the sense that it directly predicts the full API signature that contains
the API name, values of the required API arguments, relevant optional API arguments and their values.
The model chooses the API argument values to fill from
user mentioned, agent mentioned and API returned entities present in the full dialogue context
that includes the current user utterance.

We showcase the significance of our approach in reducing developer burden using the example
of a pizza ordering skill. Compared to a rule-based system where a developer would have to
code hundreds of dialogue paths to build a robust experience even for such a simple skill,
Alexa Conversations requires only a handful of seed dialogues. To evaluate our approach, we build
a movie ticket booking experience. On a test set collected
via Wizard-of-Oz~(WOZ) framework \cite{asri2017frames}, we quantify the impact of our novel dialogue simulation approach showing that it leads to
over $50\%$ improvement in action signature prediction accuracy.

\section{System Overview} \label{sec:overview}

\begin{table}[]
\scalebox{0.76}{
  \begin{tabular}{ll}
A: & \texttt{nlg: welcome()} \\
U: & ``how long is $[$ la la land $\mid$ \texttt{Movie} $\rightarrow$
\texttt{mt1} $]$'' \\
A: & \texttt{call: GetDuration(movieTitle=\$mt1)} $\rightarrow$ \texttt{d1} \\
A: & \texttt{nlg: inform\_movie\_duration(}\\
 & \qquad \texttt{duration=\$d1, movieTitle=\$mt1)} \\
U: & ``who stars in it'' \quad //anaphoric reference \\
A: & \texttt{call: GetCast(movieTitle=\$mt1)} $\rightarrow$ \texttt{gcr1} \\
A: & \texttt{nlg: inform\_movie\_cast(} \\
 & \qquad \texttt{cast=\$gcr1, movieTitle=\$mt)} \\
& ... \\
U: & ``exit'' \\
A: & \texttt{nlg: stop()}
  \end{tabular}}
  \caption{A seed dialogue with DML annotations. Note that
  variables are carried over to resolve anaphoric
  references.}\label{table:acolade_example}
\end{table}

\begin{table}[]
\centering
\scalebox{0.76}{
\begin{tabular}{l|l}
 \multicolumn{1}{c|}{\textbf{Template Name}}  & \multicolumn{1}{c}{\textbf{Template Text}}  \\
\midrule
%Inform\_PersonAge & ${personName}  is $\{age\} years old  \\
\textit{inform\_movie\_duration} & ``\$movieTitle is \$duration long''\\
\textit{inform\_movie\_cast} & ``\$cast.name was in \$movieTitle''\\
\textit{offer\_recommend\_movie} & ``Would you like a \$genre movie?'' \\
\end{tabular}}
\caption{Developer-provided system NLG responses}
\label{tab:templates}
\end{table}

In Alexa Conversations,
we follow a data-driven approach where
the developer provides seed dialogues covering the main use cases
they want to support, and annotates them in a Dialogue Markup Language (DML).
Table~\ref{table:acolade_example} shows an example of an annotated conversation.
Developers are required to provide their domain-specific APIs and custom Natural Language Generation (NLG)
responses for interacting with the user, e.g., for informing an API output
response or for requesting an API input argument as shown in Table~\ref{tab:templates}.
These APIs and system NLG responses, with their input arguments and
output values, define the domain-specific schema of entities and actions that
the dialogue system will predict. Developers also provide example user-utterances
(as templates with entity-value placeholders) which the users may use to invoke certain APIs or
to inform slot values.

To handle the wide variation of conversations a user can have with
the dialogue system, Alexa Conversations augments the developer provided
seed dialogues through a simulator.
This component takes the annotated seed dialogues as input,
and simulates different dialogue flows that achieve the same user goals
but also include common patterns such as when a user
confirms, changes, or repeats an entity or action.
Optionally, it uses crowdsourcing through Amazon Mechanical Turk (MTurk) to
enrich the natural language variations of user utterances provided by the developer.
Overall, the developer provides on the order of 10 seed dialogues
and the simulator generates on the order of 10K training dialogues with flow and
language variations.

\begin{figure}[]
  \centering
  \scalebox{1}{
  \includegraphics[width=1.0\columnwidth]{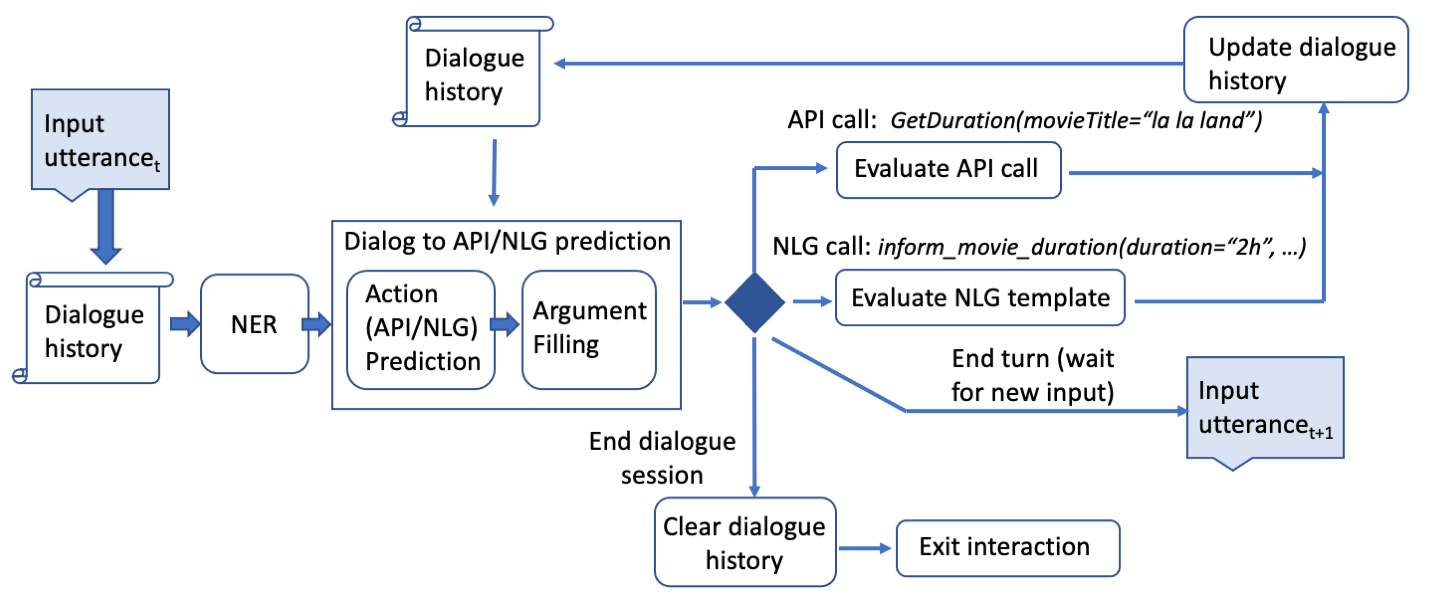}}
  \caption{High-level overview of an input utterance's path}
  \label{fig:overview}
\end{figure}

Alexa Conversations consists of three main domain-specific modeling components:
1) a Named-Entity Recognition (NER) model that tags entities in the
user utterance (e.g., “La La Land” as a MovieTitle), 2) an Action Prediction (AP) model that
predicts which API or NLG response should be executed next (e.g., \textit{GetDuration} or
\textit{inform\_movie\_duration}), and 3) an Argument Filling (AF) model that fills required (and
possibly optional) action arguments with entities (e.g.,
\textit{GetDuration(MovieTitle=``La La Land'')}). We use the entire dialogue history,
i.e., user utterances, system actions and responses, and API return values,
as input for all modeling components. In this sense,
this dialogue history is used as a generalized state representation
from which models can retrieve relevant information.
An overview of the runtime flow of a dialogue is illustrated in Figure~\ref{fig:overview}.
Each user utterance initiates a turn and is followed by NER, after which one or
more actions are predicted. These actions could be either an API or NLG call,
or a special action indicating the end of a turn or the end of dialogue.
Every new action prediction updates the dialogue
history and therefore influences future action predictions. For each API/NLG call the AF model is
called to fill in the required arguments. When \texttt{$<$end
of turn$>$} is predicted, the system waits for new user input. When
\texttt{$<$end of dialogue$>$} is predicted, the system ends the
interaction.

\section{Dialogue Simulation} \label{sec:simulator}

\begin{figure}[]
  \centering
  \includegraphics[width=1.0\columnwidth]{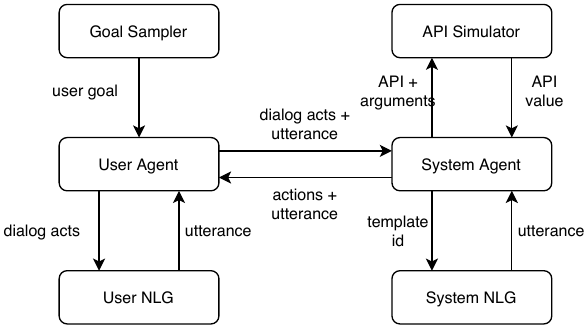}
  \caption{Simulator Architecture}
  \label{fig:simulator}
\end{figure}

We propose a novel component called simulator to generate diverse but consistent dialogues,
which can be used to train robust goal-oriented neural dialogue systems. We presented the
simulator details in \cite{lin2020dialog} and briefly provide an overview of
the overall system here. A high-level simulator architecture is illustrated in
Figure~\ref{fig:simulator}.

The simulator is structured in two distinct agents that interact turn-by-turn: the user and the
system. The user samples a fixed goal at the beginning of the conversation. We propose novel
goal-sampling techniques~\cite{lin2020dialog} to simulate variation in dialogue flows. The agents
communicate at the semantic level through dialogue acts. Having the exact information associated
with each turn allows us to define a simple heuristic system policy, whose output can be used as
supervised training labels to bootstrap models. We note that the user policy is also heuristic-based.
In each conversation, the user agent gradually reveals its goal and the system agent fulfills it by
calling APIs. The system agent simulates each API call by randomly sampling a return value
without actually calling the API and chooses an appropriate response action. Depending on the
returned API value, the chosen response is associated with dialogue acts. The system
agent gradually constructs an estimate of the user goal and makes proactive offers based on this
estimated goal. The dialogue acts generated through self-play are also used to interface between
agents and their template-based NLG model. After sampling the
dialogue acts from their policy, each agent samples the surface-form from available templates
corresponding to the dialogue acts. In addition to enriching the dialogue flows; we use
crowd-sourcing through MTurk to enrich the natural language variations
of the user utterance templates. Goal sampling and the self-play loop provide dialogue flow
variations while crowd-sourcing enriches language variations, both of which are essential
for training robust conversational models.

We introduce additional variations to dialogues during simulation for more natural conversation
generation. In goal-oriented conversations, users often change their mind during the course of
the conversation. For example, while booking a movie ticket a user may decide to purchase three
adult tickets but could eventually change their mind to book only two tickets. We used additional
heuristics to introduce such variations to conversations without any additional input requirements
from the developer.
Another important non-task-specific conversational behavior is the system's ability to suggest
an appropriate next action based on the conversation history, without requiring invocation by a
specific user utterance. We introduce proactive offers in the system policy of the simulator to
facilitate exploration of the available API functionality in a manner consistent with human
conversation.

\section{Models} \label{sec:single}
For each domain, we have three separate models: NER,
Action Prediction (AP) and Argument Filling (AF), all of which depend on
features extracted from conversation history and encoded using Dialogue Context Encoders.

\subsection{Dialogue Context Encoders}
 \label{sec:history_encoders}

\begin{figure}[]
  \centering
  \includegraphics[width=1.0\columnwidth]{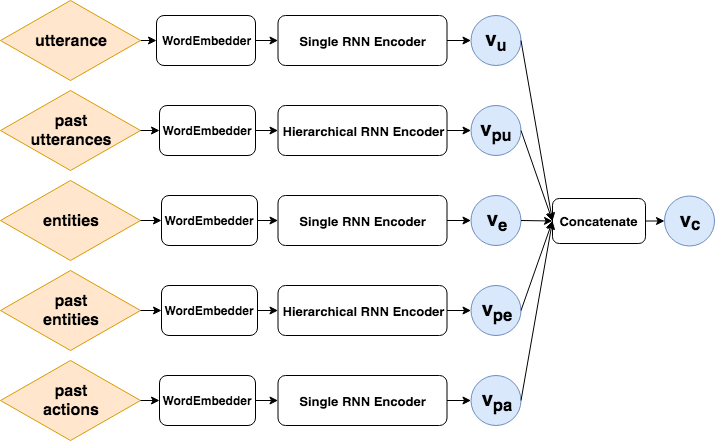}
  \caption{An example of a dialogue context encoder. Different downstream models use slightly different subsets of these features as input.}
  \label{fig:encode}
\end{figure}

Given a dialogue, we first apply feature extractors to extract both turn-level,
e.g.\ \texttt{current\_user\_utterance} and \texttt{current\_entities} (recognized by the NER
model), and dialogue-level features, e.g.\ \texttt{past\_user\_utterances}, \texttt{past\_actions}
and  \texttt{past\_entities}. We pass these extracted features through feature-specific encoders
and concatenate the feature representations to obtain the final representation
for dialogue context. For encoding turn-level features and dialogue-level features, we use single
LSTM and hierarchical LSTM architectures, respectively. For example,
for encoding \texttt{past\_user\_utterances}, we use a hierarchical LSTM, where we encode
the sequence of words with an inner LSTM and the sequence of turns with an outer
LSTM. For \texttt{past\_actions}, a single LSTM is sufficient. Figure~\ref{fig:encode}
shows an example of our dialogue context encoders. We augment the
context encoders with word and sentence embedding vectors from pre-trained
language models~\cite{peters2018deep,devlin2018bert}.

\subsection{NER}

The NER model is used to extract domain-specific entities from user utterances,
which are then consumed by downstream models. Our NER model is based on
bi-LSTM-CRF~\cite{ma_hovy_2016} model. To incorporate dialogue history, we
concatenate the encoded dialogue context to the word embedding of each token and use
it as the input to our model. To improve NER performance on entities with large
and dynamic possible values (e.g.\ movie titles, restaurant names), we also
incorporate catalogue-based features based on domain-specific catalogues of
entity values provided by the developer and values returned by APIs. Specifically,
catalogue features are computed by scanning the utterance with consecutive
windows of size $n$ tokens and detecting any exact matches of the current window
with the catalogue entries. For a domain with $K$ domain-specific catalogues,
the binary feature will be of dimension $K$, where value $1$ indicates an exact
match in the catalogue. This approach is inspired by \cite{williams2019neural}, which proposed a
generic NER approach but not specific to conversational systems.

\subsection{Action Prediction (AP)}
The goal of the Action Prediction model is to predict the next action the agent
should take, given the dialogue history. As illustrated in
Figure~\ref{fig:overview}, an action could be an API name (e.g.\ GetDuration),
a system NLG response name (e.g.\ inform\_movie\_duration) or a general system
action (e.g.\ \texttt{$<$end of turn$>$}). The model takes the dialogue context
encoding, as described in Section~\ref{sec:history_encoders} and passes
it through linear and softmax layers to output a distribution over all actions
within the domain. Our system selects $n$-best action hypotheses using a simple
binning strategy. We reject actions in the low confidence bins and if there is
no actions available in the high-confidence bin, we randomly sample an action from
the medium-confidence bin.

\begin{figure}[]
  \centering
  \includegraphics[width=0.9\columnwidth]{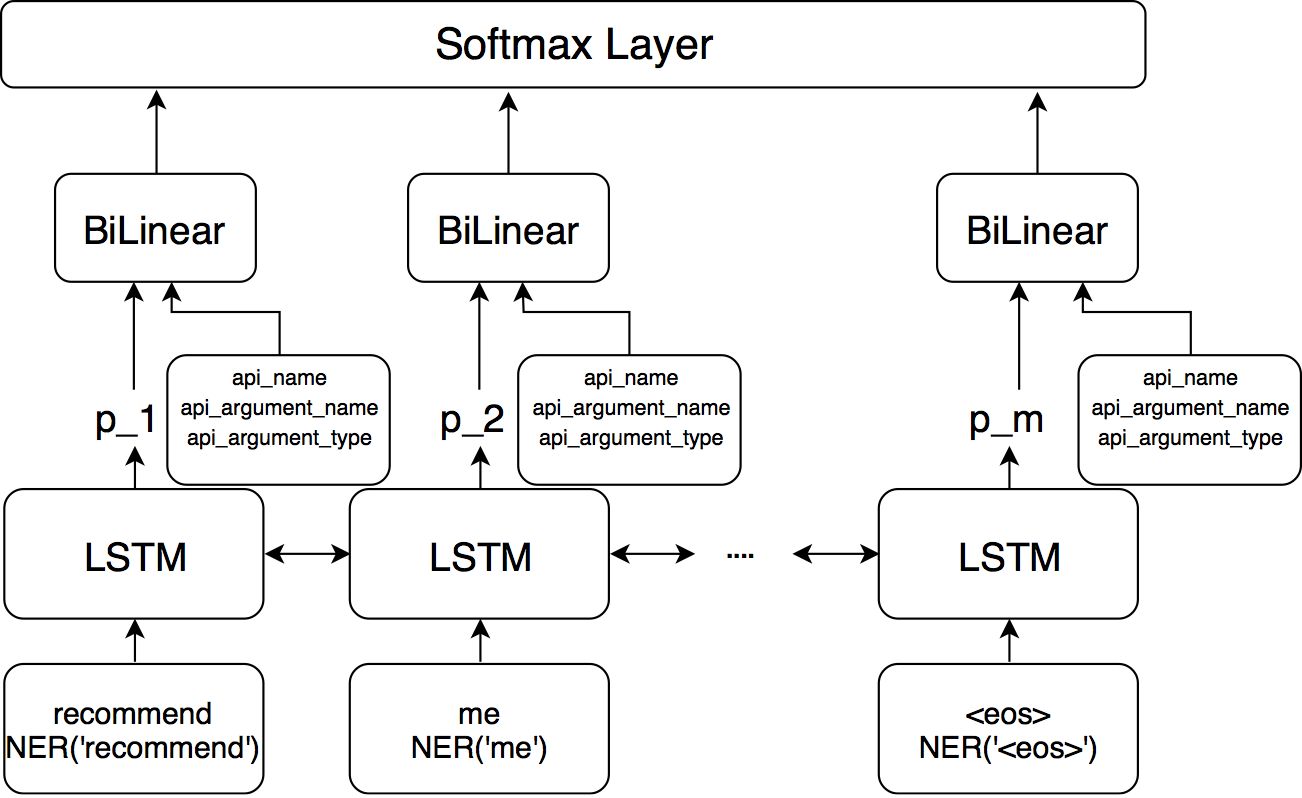}
  \caption{Argument filling model architecture}
  \label{fig:argument_filling}
\end{figure}

\subsection{Argument Filling (AF)}

The role of the Argument Filling model is to fill the arguments given a particular
action and the dialogue history. We formulate the argument filling task as a
variation of neural reading comprehension~\cite{chen2018neural} where we treat
the dialogue history as a passage to comprehend and ask machine the question
"what is the argument value of a particular action?". Specifically, for each
argument of an action and each entity mention detected by NER, our model
predicts whether to use that entity mention to fill that argument. We do this by
encoding all the entities in the dialogue history and use a pointer mechanism to
point to the entity position given a particular action and argument combination.
The overall architecture for argument filling is shown in
Figure~\ref{fig:argument_filling}. Note that a similar method to dialogue state
tracking has been proposed by \cite{gao2019dialog}. We impose constraints to
only fill arguments with entities of the correct type according to the action
schema provided by the developer. For example, we only consider \textit{Time}
entity mentions to fill arguments that accept \textit{Time} type. Finally, we
combine the argument filling result with the action prediction result to execute the next system
action. API arguments can also be optional. For these type of
arguments, we create special \texttt{optional} token and insert in the dialogue
context and train the model to point to that token when an optional API argument should
not be filled.

\section{Case Study 1: Pizzabot}
To showcase how Alexa Conversations improves the skill-building experience of a typical developer,
we discuss \textit{Pizzabot}, a pizza ordering skill. \textit{Pizzabot} is provided as one of the template
Alexa Conversations skills to help bootstrap a new developer. It allows a user to order a pizza of
various sizes, with different kinds of crusts and their choice of any number of toppings. The skill
has just one main API \textit{OrderPizza} with four slots types: \textit{toppingsList}, \textit{crust},
\textit{cheese} and \textit{size}. Even with this simple skill, ordering a two topping pizza with
a choice of say 10 toppings, 5 crusts and 5 cheese in 4 different sizes results in over hundreds of
different possible dialogue paths. With a typical rule-based dialogue manager, a developer would
need to code most, if not all, of these flows for building a robust experience. On the other hand,
with Alexa Conversations, a \textit{Pizzabot} developer needs to provide just 7 example dialogues
to support a total of 3 APIs and 6 entity types. The different variations are automatically
generated by the proposed simulator and are consumed by the models, resulting in a robust skill
with significantly lower developer effort. Alexa Conversations also provides out-of-the-box support
for features like context carryover, entity confirmation, user correction, over/under-cooperative
user, and multi-value slots without requiring the developer to write custom rules or logic, making
the skill more robust to real user interactions. See Table~\ref{tab:pizzabot} for an example
interaction with \textit{Pizzabot}.

\begin{table}[h]
    \centering
    \scalebox{1.0}{
    \begin{tabularx}{\linewidth}{X}
        \toprule
        %\textbf{Dialogue} \\ \hline
        U: I want to order a large pizza \\
        A: You wanted a large pizza, is that correct? \textit{(entity confirmation)} \\
        U: actually make it small \textit{(user correction)} \\
        A: You wanted a small pizza, is that correct? \\
        U: yes \\
        A: What would you like on your pizza? \\
        U: olives tomatoes and green peppers with thin crust and extra cheese
        \textit{(multi-value slots, over-cooperative user)} \\
        A: So far I have a small olives, tomatoes, green peppers pizza on thin crust with extra
        cheese. Is that correct? \textit{(context carryover)} \\
        U: yes \\
        A: Great! I placed your order for a total of \$5.99. \\
        \bottomrule
    \end{tabularx}}
    \caption{A seed conversation with Pizzabot}
    \label{tab:pizzabot}
\end{table}

\section{Case Study 2: Ticketbot}
\label{sec:ticket_bot}

To showcase the end-customer impact of Alexa Conversations, we built a movie ticket-booking
experience which we call \textit{Ticketbot}. \textit{Ticketbot} allows a user to browse and book
movie tickets. Users can browse currently playing movies by various search criteria like
date, time, location, theater and movie title. They can specify one or more search criteria
either within a single turn or across multiple turns. After finding their choice of movie and
theater, users can select a particular showtime, provide booking details like number of tickets
and finally confirm booking. The experience was built based on the information provided by the
developer. This is a complex experience with $10$ APIs, $28$ entity types, $10$ NLG responses
and $35$ seed dialogues all provided as an input to the system. This experience was implemented using live APIs
that were provided by the developers and thus the users were able to actually achieve their goals and complete ticket-booking transactions.

\subsection{Evaluation}

To evaluate our models, we collected data using a Wizard-of-Oz~(WOZ) framework \cite{asri2017frames}. These collected
dialogues were then annotated by a team of professional annotators using the Dialogue Markup
Language. Annotators tagged entities, API and NLG calls and unsupported requests.
This is a challenging task and we adopted various methods like inter-annotator agreement and random vetting to ensure high data
annotation quality. The test set contained $50$ dialogues with an
average length of $5.74$ turns.

We measure the F1 scores for spans of entities to evaluate NER performance.
We also measure the accuracy for action prediction (AP) and full action signature prediction (ASP).
The latter metric reflects the performance of both the AP and AF models combined:
an action signature is counted as correct when both the action and all the corresponding arguments
are predicted correctly. We compute these metrics per turn given fixed dialogue context from
previous turns, where a turn can contain one user action and multiple agent actions
(multiple api calls, nlg call, wait for user action). Turn-level ASP accuracy most closely
reflects the user experience when interacting with the skill. Overall, the system has reasonably
high turn-level action signature prediction accuracy, with relatively few failures.
We discuss some common failure patterns in \ref{err}.

We evaluate the proposed dialogue simulation method to establish the impact of this
novel component. To do so, we train models with data generated
using different simulation approaches and compare their performance on the test set.
The baseline approach, Base sampler from~\cite{lin2020dialog}, simply resamples dialogues that are identical in logical
structure to the seed dialogues. It generates no new dialogue flows but does
add language variations via sampling from developer-provided catalogs and user utterance templates.
We observe that models trained on data generated with Sec.~\ref{sec:simulator} significantly
outperform the models trained on data generated with baseline as shown in Table~\ref{table:ticketbot-results}.

\begin{table}[]
\centering
\scalebox{0.9}{
\begin{tabular}{ccc}
\toprule
    \multicolumn{1}{c}{NER Span} & \multicolumn{1}{c}{AP Relative} & \multicolumn{1}{c}{ASP Relative}   \\
      \multicolumn{1}{c}{Relative F1} & \multicolumn{1}{c}{Accuracy} & \multicolumn{1}{c}{Accuracy}   \\
    \midrule
     $+18.50\%$ & $+20.92\%$ & $+52.80\%$ \\
    \bottomrule

\end{tabular}}
\caption{Relative NER span F1-score, AP accuracy and ASP accuracy on Ticket Booking (TB) test set, averaged over $5$ runs.}
\label{table:ticketbot-results}
\vspace{-3mm}
\end{table}

\subsection{Error Analysis}
\label{err}
We conduct an error analysis of our models on the TicketBot test set to investigate performance across different tasks.
We showcase a few common error patterns in this section.

\subsubsection{NER}

We notice that NER model struggles to make correct predictions when the slot value is out of the catalogue vocabulary. As we use fixed slot catalogues during dialogue simulation, it is a difficult task for NER to generalize when real API calls return unseen values. We see that using dynamic catalogue feature significantly improves NER performance, particularly for \texttt{Movie} slot. Dynamic catalogues store entities mentioned in system's responses and thus dynamic catalogue feature provides a strong signal to NER when the user later mentions one of those entities. In addition to exact match, the feature also fires for fuzzy matches leading to higher recall without significant drop in precision. Note that, NER model is not run on system utterances; the entities are tagged by the developer in the response NLG templates. Table~\ref{tab:ticketbot_ner} shows one example in which \textit{``A Star Is Born''} is not recognized as \texttt{Movie} without this feature. Overall, by adding dynamic catalogue feature, NER improves by $+11.26\%$ on the \texttt{Movie} slot.

\begin{table}[ht]
    \centering
    \scalebox{1.0}{
    \begin{tabularx}{\linewidth}{X}
        \toprule
        U: give me [sunday $\mid$ Date] showtimes \\
        A: I found [A Star Is Born $\mid$ Movie] playing at … \textit{(mentioned movie)} \\
        U: is [captain marvel $\mid$ Movie] playing [sunday $\mid$ Date] \\
        A: I found showtimes for [Captain Marvel $\mid$ Movie] at ... \\
        U: is it playing anywhere else \\
        A: Sorry, I could not find any matching showtimes \\
        U: huh ok what are the showtimes for a star is born again \\
        ... \\
        \bottomrule
    \end{tabularx}}
    \caption{A partial dialogue from Ticketbot test set showing NER mistake}
    \label{tab:ticketbot_ner}
\end{table}

\subsubsection{ASP}

We find that Argument Filling Model makes mistakes when deciding whether or not to carryover a particular argument. In most such failures, either behavior (carryover or not carryover) is logically correct. However, the user might have a specific behavior in mind, while the model chooses the other path. For example, in Table~\ref{tab:ticketbot_reset}, AF model carryovers theater and date information, while the particular user wanted to know showtimes at all nearby theaters. As evident, this case is ambiguous as both carrying over and not carrying over the theater and date arguments is reasonable. To define the correct carryover behavior, we advise application developers to provide a few examples demonstrating the carryover behavior for each of their use cases. These examples then bias the dialogue simulator to generate data with the desired carryover behavior.

\begin{table}[ht]
    \centering
    \scalebox{1.0}{
    \begin{tabularx}{\linewidth}{X}
        \toprule
        U: What is playing at Century Cinemas tomorrow? \\
        A: \textit{call: FindMovies(theater="Century Cinemas", date="tomorrow") -> show0} \\
        A: Cold Pursuit is playing at Century Cinemas 16 at 8 PM tomorrow \\
        U: What are the showtimes for Joker? \\
        A: \textit{call: FindMovies(movie="Joker", theater="Century Cinemas", date="tomorrow") -> show1} \\
        ... \\
        \bottomrule
    \end{tabularx}}
    \caption{A partial dialogue from Ticketbot test set showing argument carryover mistake}
    \label{tab:ticketbot_reset}
\end{table}

\section{Conclusions} \label{conclusions}

We presented Alexa Conversations, a novel data-driven and data-efficient approach for
building goal-oriented conversational experiences. Our proposed system
significantly reduces developer burden while still allowing them to build robust experiences.
We envision that this system will
be used by a wide variety of developers who only need to provide seed dialogues and action
schema to build conversational experiences\footnote{iRobot (\url{https://tinyurl.com/y5pjp3xn}),
BigSky (\url{https://tinyurl.com/y2ejvd3z}) and Art Museum (\url{https://tinyurl.com/y3umpqo2}) are
some of the external skills that have already been built using Alexa Conversations}.

We expect our system to mature in the following directions in future. We aim to reduce developer
requirements for providing NLG responses by introducing a statistical NLG
system. We will also develop robust mechanisms for incorporating developer
feedback through supervised and semi-supervised methods to improve the
performance of our simulator and modeling components.

\bibliography{anthology,refs}
\bibliographystyle{acl_natbib}

\end{document}

\section{Supplementary material}
% \usepackage{graphicx}
\begin{table*}[]
\centering
\begin{tabular}{ll}
U: & What are the showtimes for Dark Phoenix on Saturday after 9PM? \\
A: & Here are the showtimes for Dark Phoenix on June 8th after 9pm. \\
U: & Get me 2­­ adult tickets for 10:35 \\
A: & Okay. 2 adult tickets for the 10:35 PM show of Dark Phoenix on Saturday, June 8th at  \\
  &  Regal Cinebarre Palace Station will be \$21. Should I book the tickets? \\
U: & Wait, how long is it? \\
A: & Dark Phoenix is 1 hour and 52 mins long. \\
U: & Actually, what are the showtimes after 7pm? \\
A: & Here are the showtimes for Dark Phoenix on June 8th after 7pm. \\
U: & Get me 2 tickets for the 7:05 show instead. \\
A: & Okay. 2 adult tickets for the 7:05 PM show of Dark Phoenix on Saturday,  \\
  & June 8th at Regal Cinebarre Palace Station will be \$21.00. Should I book the tickets? \\
U: & Yes, please. \\
A: & You will receive an email from Atom Tickets with your tickets. \\
   &  Will you be eating out near Regal Cinebarre Palace Station? \\
U: & Yes, find me a­­­ Chinese restaurant. \\
A: & Here are some Chinese ­­­­restaurants near Regal Cinebarre Palace Station \\
U: & Tell me more about Mott 32. \\
A: & Here’s more information about Mott 32. \\
U: & That looks good. Get me a table at 6:00 p.m. \\
A: & Okay. You'd like a reservation for 2 at Mott 32 at 6:00PM. Should I book it? \\
U: & Sure. \\
A: & You will receive an email from Open Table with the details. Would you like a cab to Mott 32? \\
U: & Yes, get me an Uber. \\
A: & Sure. I'll get you an Uber X for 2 from home to Mott 32, which will cost about \$8. Is that okay? \\
U: & Sounds good. \\
A: & Your Uber X will pick you up at 5:42PM. Anything else? \\
U: & Show me the trailer. \\
A: & Now playing the trailer for Dark Phoenix.
\end{tabular}%
\caption{One example multi-domain interaction between user and agent for the night out experience. }
\label{tab:night_out}
\end{table*}